 \documentclass[tablecaption=bottom,wcp]{jmlr} 

\usepackage{eurosym,array,ragged2e,booktabs,caption}
\usepackage{booktabs}
 \usepackage[utf8]{inputenc}
\usepackage{multirow} 
\usepackage[load-configurations=version-1]{siunitx} 
\usepackage{graphicx}
\newcommand{\indep}{\rotatebox[origin=c]{90}{$\models$}} 

\theorembodyfont{\upshape}
\theoremheaderfont{\scshape}
\theorempostheader{:}
\theoremsep{\newline}

\jmlrproceedings{AABI 2019}{Mbuvha Boulkaibet and Marwala}

\title[An Automatic Relevance Determination Knockoff Filter]{An Automatic Relevance Determination Prior Bayesian Neural Network for Controlled Variable Selection}

 \author{\Name{Rendani Mbuvha} \Email{mbuvha.rendani@wits.ac.za}\\
 \Name{Illyes Boulkaibet} \Email{ilyesb@uj.ac.za }\\
 \Name{Tshilidzi Marwala}\Email{tmarwala@uj.ac.za}\\
  \addr University of Johannesburg, South Africa}

\begin{document}

\maketitle

\begin{abstract}
We present an Automatic Relevance Determination prior Bayesian Neural Network(BNN-ARD) weight l2-norm measure as a feature importance statistic for the model-x knockoff filter. We show on both simulated data and the Norwegian wind farm dataset that the proposed feature importance statistic yields statistically significant improvements relative to similar feature importance measures in both variable selection power and predictive performance on a real world dataset.
\end{abstract}


\section{Introduction}
\label{sec:intro}


Central to many machine learning tasks is a feature selection problem where candidate features are considered such that they maximise the quality and strength of the signal they give about the dynamics of the system or process being modelled\citep{lu2018deeppink,candes2018panning}. This will typically require expert judgement or computationally expensive iterative variable selection methods\citep{candes2018panning}. 

The knockoff filter of \cite{candes2018panning} provides a framework for performing variable selection while controlling the false discovery rate (FDR) of null features. Model-X knockoffs work by creating 'knockoff' variables that imitate the dependency structure within the original candidate feature space but critically with no relationship with the target variable \citep{barber2015controlling}. These knockoff features can thus be used as variable selection controls for their corresponding candidate features. If a candidate feature on the basis of a specified feature importance metric is less important than its corresponding knockoff then it follows that the candidate feature is irrelevant. Various feature importance metrics have been suggested for determining the relative importance between candidate features and their knockoffs. These have included LASSO coefficients\citep{barber2015controlling,candes2018panning}, filtering neural network layers \citep{lu2018deeppink} and permutation based methods\citep{gimenez2018knockoffs}. 

In this paper we integrate weight l2-norm feature importance statistics derived  from Automatic Relevance Determination(ARD) prior Bayesian Neural Networks(BNNs) into the knockoff filter framework. We show that the proposed feature importance statistics obtain the desired FDR control and display outperformance in statistical power to other nonlinear feature importance measures.

\section{Background}\label{knocffs}

The modelling setting we consider is one where we have independent and identically distributed observations of candidate feature vectors $(X_{i1}...,X_{ip})\in {R}^p$  and a corresponding scalar $Y_i$, $i=1,....,n$. We assume that Y depends on a subset $\mathcal{H}_0$ of the complete set of candidate features $\mathcal{S}_0$ and  is conditionally independent of the features in complement of $\mathcal{H}_0$ given the features in $\mathcal{H}_0$.

The controlled variable selection problem is one that aims to discover as many of the features in $\mathcal{H}_0$ as possible while controlling the FDR which is defined as follows for a set of selected  features $\hat{\mathcal{S}}$\citep{gimenez2018knockoffs}:
\begin{equation}
    \text{FDR} = \mathbb{E}[\text{FDP}] \quad \text{with} \quad \text{FDP} = \frac{|\hat{\mathcal{S}}	\cap\mathcal{H}_0|}{|\mathcal{S}\vee 1|}
\end{equation}

The recently proposed model-x knockoffs filter \citep{candes2018panning} has been widely employed as a mechanism for achieving FDR control below some desired threshold $q$.
\begin{definition}[Model-X Knockoffs features\citep{candes2018panning}]\label{def:sample}
A model-x knockoff for a $p$-dimensional random variable vector ${X}$ is a $\textit{p}$-dimensional random variable vector $\widetilde{X}$  that satisfies the following properties for any subset $\mathcal{S}\in (1,...,p)$:
\begin{enumerate}

    \item $\big(X,\widetilde{X}\big)_{\text{swap(s)}}\,{\buildrel d \over =}\big(X,\widetilde{X}\big)$ where $\text{swap(s)}$ is an operation swapping corresponding entries of $X$ and $\widetilde{X}$ for each $j\in \mathcal{S}$
    \item $\widetilde{X}\indep Y|X$ i.e. the response Y is conditionally independent of the knockoff features given the candidate features.
\end{enumerate}

\end{definition}

In special cases where features are Gaussian $X \sim \mathcal{N}\big(0,\Sigma\big)$, model-x knockoffs can be constructed by solving constrained optimisation problems for matching second order moments.

Once the model-x knockoffs are constructed feature importance statistics $Z_j$ and $\widetilde{Z}_j$ for   $X_j$ and $\widetilde{X}_j$ with $j\in (1,...,p)$. $W_j=f(Z_j,\widetilde{Z}_j)$ ard then defined as distance metric between feature importance of a candidate feature  $X_j$ and its knockoff $\widetilde{X}_j$ this is typically the l1-norm. 
Given the relative feature importance statistics $W_j$, the $|W_j|$ are sorted in descending order and we then select the features where $|w_j|$ exceed the threshold $T$ which is defined for a target FDR $q$ as  : 
\begin{equation}\label{thresh}
    T = min\bigg\{t\in \mathcal{W},\frac{1+|\{j:W_j\leq-t\}|}{1\vee|\{j:W_j\geq-t\}|}\leq q\bigg\}
\end{equation}

\section{Automatic Relevance Determination and Weight l2-norm Feature importance Statistics}\label{ARD}
An ARD prior for a BNN is one where weights associated with each input feature connection to the first hidden layer belong to a distinct class and have unique precision parameters $\alpha_c$ for each input. The posterior distribution will then be as follows\citep{mackay1995probable}:

\begin{equation} 
P(w|D,\alpha,\beta)=\frac{1}{Z(\mathbf{\alpha},\beta)}\exp\bigg(\beta E_D+\sum_{c}{\alpha_c} E_{W_C}\bigg)
\end{equation}
Where $\alpha_c E_W$ is the kernel of the prior distribution on the weights , $\beta E_D$ is the kernel of the data likelihood and $Z(\mathbf{\alpha},\beta)$ is the normalisation constant. 
The precision hyperparameters for each group of weights are estimated by evidence maximisation or Markov chain Monte Carlo\citep{mackay1995probable,neal2012bayesian}. Irrelevant features will have high values of the regularisation parameter meaning that their weights will be forced to decay to values close to zero.

In order to manage the scaling of the regularisation parameters we then consider the l2-norm of the resulting weights as a measure of relative feature importance. We infer the importance of a feature $I(f)$ using the l2-norm all the weights connecting a specific input to the first hidden layer units i.e.
\begin{equation}
 \textit{I}(f)=\sum_{w\in f}{w_{ij}^2}
\end{equation}

We incorporate this feature importance measure in the  model-x knockoff filter by setting $I(f)$ =  $Z_f$.

\section{Experiments}
\subsection{Simulation Studies}
We use simulated data to compare the variable selection performance of our proposed BNN-ARD weight l2-norm feature importance statistic with other non-linear feature importance statistics. We simulate non-linear data from the data model; $ Y = \frac{\big(\mathbf{X}^T\mathbf{\beta}\big)^3}{2} + \varepsilon$. Where Y is a vector of the response, $\mathbf{\beta}$ is a p dimensional vector of coefficients, $\mathbf{X}\in \mathbb{R}^{n\times p}$ is a feature matrix sampled independently from $X \sim \mathcal{N}\big(0,\Sigma\big)$ where elements of the co-variance matrix are set to  $\big(\Sigma\big)_{jk}=(\rho^{|j-k|})_{jk}$ with $j,k\in[1,p]$ and $\rho=0.5$.   

We randomly set the amplitudes of a subset of relevant features  $\mathcal{H}_0$ of 10 randomly selected elements to a value of $3.5$. The total number of candidate features in  $\mathcal{S}_0$, $p$ was set to 100. We the compare knockoff filters with the following feature importance statistics: BNN-ARD weight l2-norm; Multi-layer Perceptron(MLP) weight l2-norm ;Random forest(RF) mean decrease in accuracy\citep{breiman2001random}. We repeat the experiment with 100 random initialisations and report on the mean power and FDR as the target FDR varies.

\subsection{Real Data Experiments}
We further assess the relative performance of the knockoff filters based on the various feature importance statistics on the Norwegian wind farm dataset.  The dataset consists of 7384 records covering the period from January 2014 to December 2016\citep{mbuvha2017bayesian}.  The eleven candidate features include the wind farm online capacity, one and two hour lagged historical power production values as well as numerical weather prediction(NWP) estimates of humidity, temperature and wind speed. 

We compare the performance of the various feature importance statistics based on the testing set Root Mean Square(RMSE) of an MLP trained on features selected by each of the respective feature importance statistics.

\section{Results and Discussion}
\begin{figure}[ht]

  \caption{Left,Plot of the mean power of the respective feature importance statistics after 100 simulations across various target FDRs. Right, Plot of the mean empirical FDR of the respective feature importance statistics after 100 simulations across various target FDRs. }
  \includegraphics[scale=0.2]{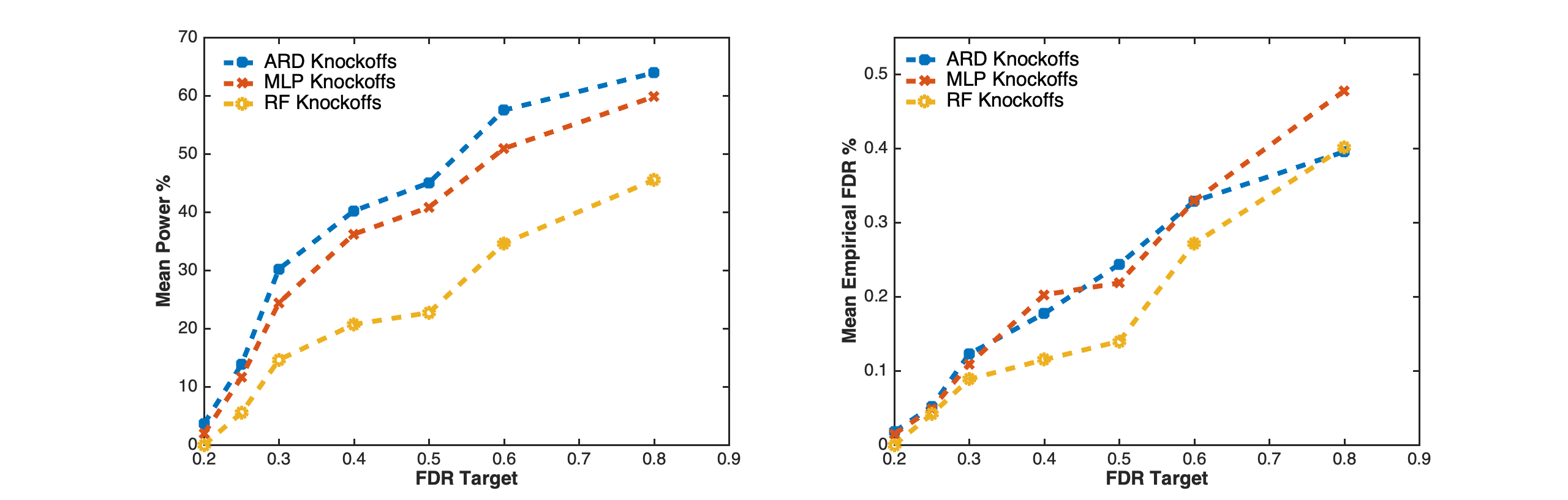}
  \label{fig:sim}
\end{figure}

Figure \ref{fig:sim} shows the mean power and empirical FDR across various target FDR levels of the 100 simulations. The results show that the BNN-ARD l2-norm knockoff filter displays persistently higher power in recovering non-null features, relative to the MLP l2-norm  knockoff filter and the RF knockoff filter respectively. The corresponding empirical FDR plot is  relatively linear below the 45 degree line meaning that all knockoff filters maintained the theoretical Target FDR. A non-parametric Kruskal Wallis\citep{mckight2010kruskal} test for statistical significance of the differences in power yields a strongly significant p-value of $9.45e-18$. A further Bonferroni test\citep{bonfurronie} on the pair wise differences in power show statistical significance at a 10 percent level between the BNN-ARD l2-norm and MLP l2-norm filters and at the 5 percent level for both neural network based filters relative to the RF filter.   

Table \ref{tab:wind} shows the mean testing RMSE on the Norwegian wind farm dataset. The results show that the mean testing RMSE based in features selected by the BNN-ARD l2-norm filter is lower than that of the MLP l2-norm filter with statistical significance at FDR target levels greater than 0.25. The features selected by both filters at their respective minimum testing RMSE displayed an 88\% overlap. Features such as the third order lag in power production, the second order lag in plant availability and the NWP temperature forecast were irrelevant. The RF filter results are unavailable as the filter yielded empty sets for FDR targets below 0.5. These results show congruence with work previously done on the same dataset by \cite{mbuvha2017bayesian}.
 
\begin{table}[ht]
\begin{center}
\begin{tabular}{c c c c c c}
    \hline
    Feature Importance Statistic  & \multicolumn{5}{c}{FDR Target}\\
    \cmidrule{2-6}
    & 0.2 & 0.25 & 0.3 & 0.4 & 0.5   \\ 
    \hline
    MLP-l2 Norm & 7257.59	& 6744.63 &	7427.79 &	4398.25	& 3229.72 \\
    ARD-l2 Norm & 7325.57 &	6124.28 &	5173.02 & 3717.99 & 2991.72	\\
    
    \hline
\end{tabular}
\end{center}
\caption{The mean RMSE based on the different knockoff feature importance statistics after 30 random initialisation across various target FDRs on the Norwegian wind farm dataset}
\label{tab:wind}
\end{table}
 

\bibliography{references}

\appendix





\end{document}